\documentclass[runningheads]{llncs}
\usepackage{graphicx}
\usepackage{subcaption}
\usepackage{multirow}
\usepackage{tikz}
\usepackage{algorithm}
\usepackage{algpseudocode}
\usepackage{pgfplots}

\usepackage{amsmath}
\usepackage{amssymb}
\usepackage{bm}
\usepackage{booktabs}

\usepackage{wrapfig}

\usepackage{eso-pic}
\usepackage{xcolor}

\usepackage{xspace}

\begin{document}
\title{Boosting Adversarial Transferability via \\ Commonality-Oriented Gradient Optimization}
\titlerunning{Boosting Adversarial Transferability} 

\author{Yanting Gao\inst{1} \and Yepeng Liu\inst{2} \and
Junming Liu\inst{1} \and Qi Zhang\inst{1} \and Hongyun Zhang\inst{1} \and Duoqian Miao\inst{1} \and Cairong Zhao\inst{1}}
\authorrunning{Gao et al.}

\institute{Tongji University \and University of Florida}

\AddToShipoutPicture*{
  \put(120,80){\footnotesize\color{black}Preprint.}
}
\maketitle

\begin{abstract}
Exploring effective and transferable adversarial examples is vital for understanding the characteristics and mechanisms of Vision Transformers (ViTs). However, adversarial examples generated from surrogate models often exhibit weak transferability in black-box settings due to overfitting. Existing methods improve transferability by diversifying perturbation inputs or applying uniform gradient regularization within surrogate models, yet they have not fully leveraged the shared and unique features of surrogate models trained on the same task, leading to suboptimal transfer performance. Therefore, enhancing perturbations of common information shared by surrogate models and suppressing those tied to individual characteristics offers an effective way to improve transferability. Accordingly, we propose a commonality-oriented gradient optimization strategy (\textbf{COGO}) consisting of two components: Commonality Enhancement (CE) and Individuality Suppression (IS). CE perturbs the mid-to-low frequency regions, leveraging the fact that ViTs trained on the same dataset tend to rely more on mid-to-low frequency information for classification. IS employs adaptive thresholds to evaluate the correlation between backpropagated gradients and model individuality, assigning weights to gradients accordingly. Extensive experiments demonstrate that \textbf{COGO} significantly improves the transfer success rates of adversarial attacks, outperforming current state-of-the-art methods.

\end{abstract}

\section{Introduction}

Vision Transformer (ViT)~\cite{dosovitskiy2020image} and its variants~\cite{liu2021swin,raghu2021vision,touvron2021training,wu2020visual,zhang2024mg} have demonstrated strong performance in computer vision tasks by effectively capturing long-range dependencies and contextual information. However, ViTs are highly vulnerable to adversarial attacks: even small and well-designed perturbations can lead to severe misclassifications, seriously compromising their reliability in safety-critical applications~\cite{goodfellow2014explaining,liu2024image,madry2017towards,shao2021adversarial}. Exploring and developing universal and effective adversarial attack strategies is essential for improving model robustness.  In this context, transfer-based adversarial attacks on ViTs have drawn increasing attention, as they do not require the internal structure or setting of the target model~\cite{goodfellow2014explaining,inkawhich2020transferable}, making them more applicable to real-world scenarios~\cite{liu2016delving,papernot2017practical}.

Transfer-based adversarial examples generated by surrogate models often encode excessive surrogate-specific information, leading to poor generalization to unseen target models, a phenomenon commonly referred to as adversarial overfitting~\cite{hamdi2020advpc,moosavi2017universal,tramer2017ensemble,wu2020skip}. To alleviate this issue, gradient ascent is commonly enhanced with input diversity and gradient regularization. For instance, High-Frequency Adaptation (HFA)~\cite{zhuenhancing} introduces input diversity by amplifying high-frequency components through random noise. However, it neglects the fact that surrogate models strongly depend on mid-to-low frequency information for classification. Meanwhile, methods like token gradient regularization (TGR)~\cite{zhang2023transferable} and Gradient Normalization Scaling (GNS)~\cite{zhuenhancing} regularize gradients more strategically: TGR suppresses extreme gradients, and GNS enhances moderate ones. Despite their effectiveness, these methods uniformly adjust gradients based solely on statistical properties, without considering the correlation between gradients and model-specific characteristics. In other words, if gradient regularization is guided by the objective of weakening surrogate-specific features, it can enhance transferability; otherwise, it remains constrained by surrogate-specific biases, ultimately compromising black-box performance.

Building on this insight, we turn our attention to leveraging features that are commonly shared across diverse ViTs. Although prior studies have shown that ViTs are vulnerable to high-frequency perturbations, this vulnerability reflects architectural sensitivity rather than common decision behavior, since high-frequency responses differ significantly across models. In contrast, ViTs trained on the same dataset tend to rely consistently on mid-to-low frequency components, which encode shape and semantic structures, as the primary basis for classification~\cite{kim2024exploring,shao2021adversarial,zhang2024mlip}. These components therefore serve as a better representation of model commonality. However, identifying such commonality directly in black-box scenarios is difficult due to the inaccessibility of target gradients. To address this, we propose applying frequency-aware perturbations to the mid-to-low frequency components of input images, where the perturbation strength is adapted based on their semantic importance. This encourages the generated adversarial examples to exploit features that are commonly used across different ViTs, thereby achieving \textbf{commonality enhancement}.

\begin{figure}[t]
    \centering
    \setlength{\tabcolsep}{0pt}
    \begin{tabular*}{\textwidth}{@{\extracolsep{\fill}}cccc@{}}
        \begin{subfigure}{0.2\textwidth}
            \centering
            \includegraphics[width=\textwidth]{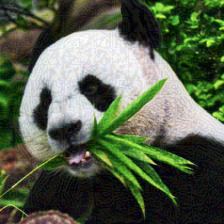}
            \caption{}
            \label{fig:adv_example_vit}
        \end{subfigure}
        &
        \begin{subfigure}{0.24\textwidth}
            \centering
            \includegraphics[width=\textwidth]{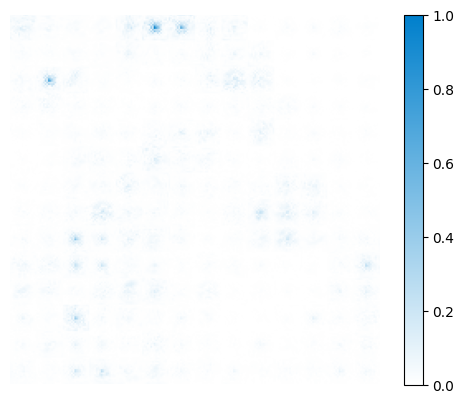}
            \caption{}
            \label{fig:gradient_sensitivity_vit}
        \end{subfigure}
        &
        \begin{subfigure}{0.2\textwidth}
            \centering
            \includegraphics[width=\textwidth]{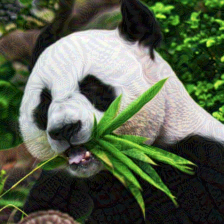}
            \caption{}
            \label{fig:adv_example_visformer}
        \end{subfigure}
        &
        \begin{subfigure}{0.24\textwidth}
            \centering
            \includegraphics[width=\textwidth]{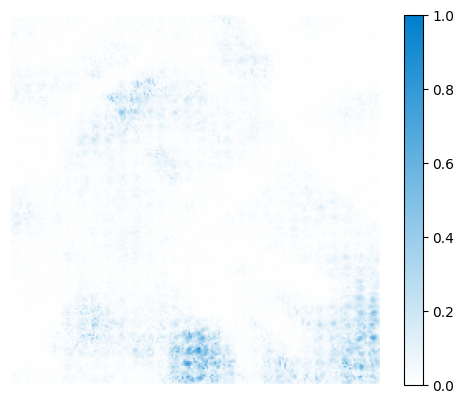}
            \caption{}
            \label{fig:gradient_sensitivity_visformer}
        \end{subfigure}
    \end{tabular*}
    
    \caption{A comparison of adversarial examples and gradient sensitivity maps generated for the same input image across different models (ViT-base and Visformer). Specifically, (a) and (c) show the adversarial examples generated by ViT-base and Visformer, while (b) and (d) show their corresponding gradient sensitivity maps.}
    \label{fig:combined_adv_grad}
\end{figure}

From a complementary perspective, suppressing the individuality of surrogate models also contributes to optimizing for commonality. Due to architectural differences, various ViTs extract features in distinct ways and thus learn heterogeneous representations even when trained on the same dataset~\cite{han2021transformer,touvron2021training}. This leads to varying gradient sensitivities, resulting in different adversarial perturbations across models, as illustrated in Figure~\ref{fig:combined_adv_grad}. When perturbations are generated based on surrogate gradients, some components correspond to shared decision patterns and support transferability, while others reflect surrogate-specific biases and hinder it. Therefore, explicitly reducing the influence of these individual-specific gradients is crucial for promoting transfer-friendly perturbations through \textbf{individuality suppression}.

In this paper, we propose \textbf{Commonality-Oriented Gradient Optimization (COGO)}, a novel strategy to enhance adversarial transferability by focusing on shared decision patterns of ViTs trained on the same dataset and task, while considering structural differences that cause gradient inconsistencies. \textbf{COGO} consists of two key components: \textbf{Commonality Enhancement (CE)} and \textbf{Individuality Suppression (IS)}. During forward propagation, \textbf{CE} amplifies mid-to-low frequency components by applying frequency energy enhancement via Discrete Cosine Transform (DCT). In backward propagation, \textbf{IS} employs adaptive thresholds to identify and suppress gradients linked to surrogate-specific features, preserving those aligned with shared model characteristics to guide perturbation updates. By simultaneously emphasizing shared semantic features and mitigating surrogate-specific biases, \textbf{COGO} offers a principled approach to improve adversarial transferability across various ViTs.

Experiments demonstrate that \textbf{\textit{COGO}} significantly improves the transferable attack success rate, outperforming the advanced method GNS-HFA by up to 16.1\%. In summary, our contributions are as follows:

\begin{itemize}
    \item We adopt a commonality-oriented optimization that accounts for the relationship between gradients and model characteristics, moving beyond traditional uniform gradient adjustment methods.
    \item We propose \textbf{\textit{COGO}}, a novel strategy combining \textbf{Commonality Enhancement (CE)} and \textbf{Individuality Suppression (IS)}. It leverages frequency-domain energy enhancement and an adaptive threshold mechanism to strengthen model commonality while suppressing surrogate-specific individuality.
    \item \textbf{\textit{COGO}} significantly outperforms state-of-the-art methods, including GNS-HFA and ATT, demonstrating its effectiveness.

\end{itemize}

\section{Related Work}
\label{sec:rw}

\subsection{Vision Transformers}
\label{sec:vit_optimization}
Transformers have played a pivotal role in natural language processing, giving rise to numerous architectural variants~\cite{fu2025dual,xu2024skinformer,xu2024mrftrans}. At the same time, they serve as a cornerstone for the development of large language models, where their effectiveness is indispensable~\cite{lin2023advances,Liu_2025_VaLiK}. However, the widespread application of these LLM systems, enabled by Transformers, has also exposed various security issues~\cite{he2024theoretically,liu2024adaptive,liu2025dataset,lou2023trojtext,xue2023trojllm}. Beyond NLP, Transformers have also driven progress in core computer vision tasks~\cite{lin2023advances,wang2017adversarial,wang2023accurate,wu2025segment,xu2024skinformer} and multimodal learning tasks~\cite{shen2025efficient,wan2025srpo,wan2025d2o,xu2024mrftrans,zhang2025enhancing}, providing effective technical solutions for these fields, further expanding the practical value of Transformer-based methods~\cite{fu2024fast,fu2025dual,zhang2025cae}. ViTs~\cite{dosovitskiy2020image}, adapted from the self-attention mechanism in Transformers, have popularized computer vision~\cite{wang2023accurate,wu2025segment}. Since their introduction, numerous variants have emerged, including Visformer~\cite{chen2021visformer}, DeiT~\cite{touvron2021training}, PiT~\cite{heo2021rethinking}, CaiT~\cite{touvron2021going}, TNT~\cite{han2021transformer} and MG-ViT~\cite{zhang2024mg}. However, similar to Transformers in natural language processing, ViTs have been found vulnerable to adversarial attacks, which motivates our research in this direction.

In this work, we categorize ViT variants based on their dominant design priorities. One group focuses on computational efficiency by simplifying attention operations or integrating convolutions to reduce inference cost, as seen in Visformer~\cite{chen2021visformer} and PiT~\cite{heo2021rethinking}. The other group emphasizes data efficiency by improving representation capacity and generalization, often through mechanisms such as knowledge distillation~\cite{wu2025segment} or deeper attention stacks, as in DeiT~\cite{touvron2021training} and CaiT~\cite{touvron2021going}. It is important to note that this categorization is not mutually exclusive—many models exhibit traits from both categories. However, we analyze and group them based on their most salient characteristics to guide the design of our method. Although these models are typically trained on the same datasets and tasks and tend to form shared decision patterns, their architectural differences lead to diverse gradient behaviors. This heterogeneity can reduce adversarial transferability, motivating the need to craft perturbations that emphasize shared patterns while suppressing surrogate-specific biases.

\subsection{Transfer-Based Black-Box Adversarial Attack}

In transfer-based black-box attacks, adversarial examples generated on a surrogate model are designed to exploit cross-model vulnerabilities and transfer to attack unknown target models~\cite{liu2016delving}. Numerous methods have been developed to improve adversarial transferability.

Methods targeting CNNs as surrogate models have a long history. Early methods like BIM~\cite{kurakin2018adversarial}, PGD~\cite{madry2017towards}, and MIM~\cite{dong2018boosting} focused on stabilizing gradient directions, while DIM~\cite{xie2019improving} and TIM~\cite{dong2019evading} improved transferability via input transformations. Recently, SSA~\cite{long2022frequency} explored frequency-domain vulnerabilities comprehensively, but did not design frequency-aware perturbations specifically to enhance transferability. Additionally, HFA~\cite{zhuenhancing}, inspired by SSA, only adapts high-frequency components, neglecting mid-to-low frequencies. These limitations lead to suboptimal results on ViTs, motivating our method to leverage a broader frequency spectrum tailored for ViTs to improve transferability.

\begin{figure*}[ht]
    \centering
    \includegraphics[width=1\textwidth]{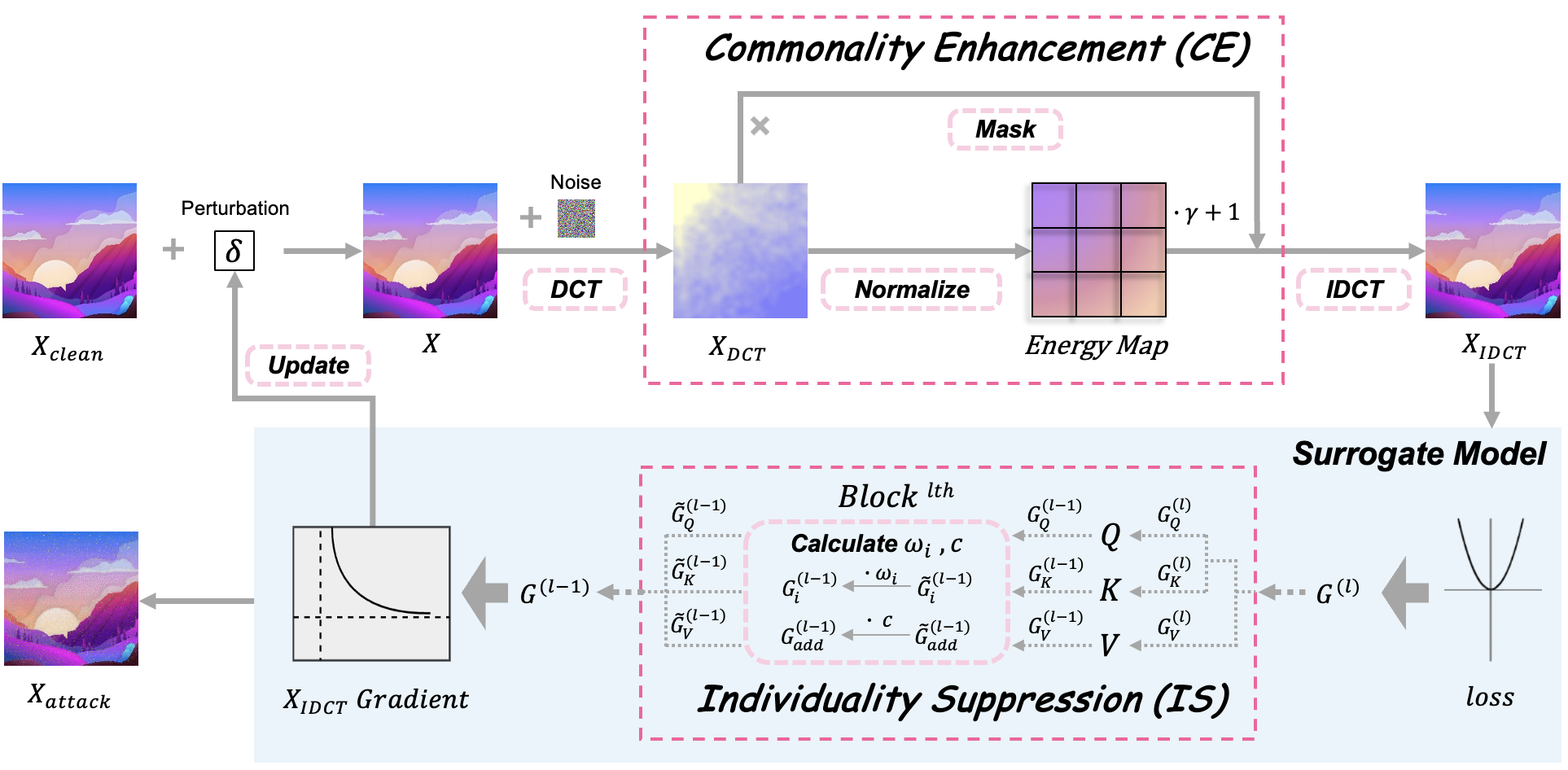}
    \caption{
An iteration of adversarial example generation under the COGO strategy, which enhances mid-to-low frequency components via Commonality Enhancement (CE) and suppresses surrogate-specific gradients via Individuality Suppression (IS) to improve transferability.
}
    \label{fig:flowchart}
\end{figure*}

Recent methods such as TGR~\cite{zhang2023transferable} and GNS-HFA~\cite{zhuenhancing} have advanced adversarial transferability on ViTs by mitigating gradient overfitting. TGR reduces the variance of gradients within the surrogate model by suppressing extreme gradient values, hypothesizing that these outliers drive overfitting. This effectively decreases internal gradient variance. Conversely, GNS-HFA posits that moderate gradients are more crucial for transferability, regularizing gradients towards a Gaussian distribution and amplifying those below a threshold to promote uniformity. It also employs high-frequency adaptation to enhance input diversity and further improve transferability.

Nevertheless, both approaches focus solely on homogenizing the gradient distribution within the surrogate model, without explicitly perturbing the shared decision patterns across different architectures. Consequently, their adversarial examples achieve only suboptimal transferability, as these methods inevitably incorporate some ineffective gradient updates. Furthermore, experiments show a notable performance drop when transferring attacks to CNN-based targets, reflecting the intrinsic gap between ViTs and CNNs in adversarial robustness.

\section{Method}
\label{sec:method}

\subsection{Overview}
We provide an overview of the COGO strategy in Figure~\ref{fig:flowchart}, illustrating the process of a single iteration.

Starting with a zero-initialized perturbation added to the input, we apply a Discrete Cosine Transform (DCT) to project it into the frequency domain. There, the Commonality Enhancement (CE) module selectively amplifies mid-to-low frequency components based on an energy map, guiding the perturbation toward shared decision patterns across architectures. The modified frequency representation is then converted back to the spatial domain via Inverse DCT (IDCT) and passed through the surrogate model.

During backpropagation, the Individuality Suppression (IS) module adaptively suppresses surrogate-specific gradients using a thresholding mechanism, reducing overfitting and promoting transferable directions. This process iterates until convergence, producing the final adversarial example.

\begin{figure}[t]
    \centering
    \hspace*{\fill} 
    \begin{subfigure}{0.29\textwidth}
        \centering
        \includegraphics[width=\textwidth]{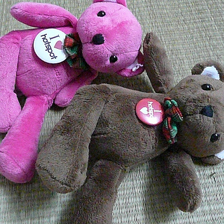}
        \caption{}
        \label{fig:bear}
    \end{subfigure}
    \hspace{0.04\textwidth} 
    \begin{subfigure}{0.36\textwidth}
        \centering
        \includegraphics[width=\textwidth]{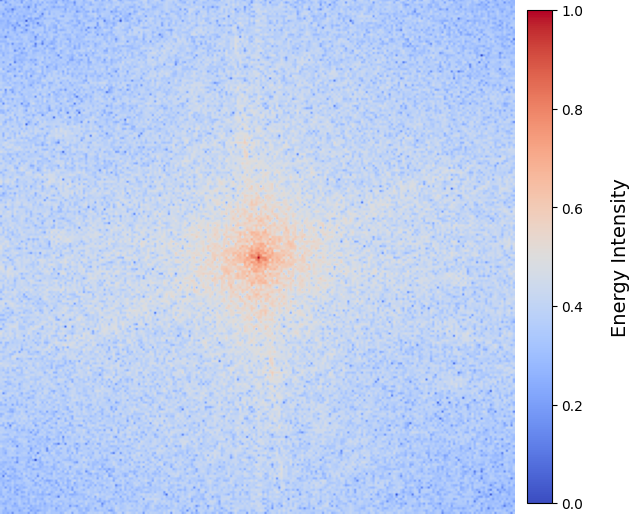}
        \caption{}
        \label{fig:energy}
    \end{subfigure}
    \hspace*{\fill} 
    \caption{(a) shows the original input, while (b) illustrates the energy intensity distribution in the frequency domain. It can be observed that the low-frequency components, typically located near the center of the spectrum, exhibit stronger energy compared to the high-frequency components.}
    \label{fig:energy_map}
\end{figure}

\subsection{Commonality Enhancement}

In the forward process of adversarial example generation, most existing methods focus on high-frequency perturbations~\cite{zhuenhancing}. However, recent studies~\cite{zhang2024mlip} show that ViTs rely more on mid-to-low frequency components, which encode essential structural and semantic information. As illustrated in Figure~\ref{fig:energy_map}, these components dominate the energy spectrum, which are important in feature representation. Enhancing them not only reinforces ViT decision signals but also promotes gradients aligned with shared patterns, thereby improving adversarial transferability.

To leverage this property, our CE module adaptively enhances mid-to-low frequency components using the energy distribution of each input sample. Starting from a clean input \(X_{\mathsf{clean}} \in \mathbb{R}^N\), we add the current perturbation \(\delta \in \mathbb{R}^N\) (initialized as zero) to obtain:
\begin{align}
X = X_{\mathsf{clean}} + \delta.
\end{align}
Gaussian noise \(\epsilon \sim \mathcal{N}(0, I_N)\) is added to increase frequency diversity, and we apply the Discrete Cosine Transform (DCT) to map the perturbed sample to the frequency domain:
\begin{align}
X_{\mathsf{DCT}} &= \mathsf{DCT}(X + \epsilon), \\
E(X_{\mathsf{DCT}}) &= \mathsf{Normalize}(|X_{\mathsf{DCT}}|),
\end{align}
where \(E(X_{\mathsf{DCT}})\) represents the normalized energy map. We then enhance the frequency representation in energy-dominant regions:
\begin{align}
X_{\mathsf{DCT}}' &= X_{\mathsf{DCT}} \cdot \left( 1 + \gamma \cdot E(X_{\mathsf{DCT}}) \right),
\label{eq:dct}
\end{align}
with \(\gamma\) controlling the enhancement strength. Finally, the enhanced frequency sample is converted back to the spatial domain using IDCT. We apply a spatial mask \(M\), inherited from HFA~\cite{zhuenhancing}, to introduce additional perturbation diversity:
\begin{align}
X_{\mathsf{IDCT}} &= \mathsf{IDCT}(X_{\mathsf{DCT}}' \cdot M),
\end{align}
and feed \(X_{\mathsf{IDCT}}\) into the surrogate model.

\subsection{Individuality Suppression}
\paragraph{Gradient Suppression Location.}  
We apply Information Suppression (IS) to the \(qkv\) module, as it directly controls token-to-token interactions in ViTs and plays a central role in capturing both common and model-specific patterns. Compared to other modules like the Feed-Forward Network (FFN), which processes each token independently, the \(qkv\) module enables global information flow via attention. Suppressing gradients here thus has a broader impact, making it more effective for reducing surrogate-specific biases. Empirical comparisons with alternative locations are provided in Appendix~\ref{sec:details}.

\paragraph{Two types of features requiring gradient suppression.}
Inspired by the optimization approaches of ViT variants in Section~\ref{sec:vit_optimization}, we identify two types of features requiring gradient suppression: (1) those associated with redundant representations in structurally optimized ViTs, and (2) those linked to additional knowledge in data-efficient variants. The suppression algorithm is shown in Appendix~\ref{alg:fga}.

\paragraph{IS for Redundant Features.} 
For structurally optimized ViTs, we address the issue of feature redundancy, where multiple channels capture overlapping information, leading to imbalanced gradient updates that overemphasize certain feature directions and thus reduce adversarial transferability. To quantify redundancy, we utilize widely accepted metrics Mutual Information (MI) and Pearson Correlation (PC) between gradient channel pairs, which effectively measure shared information and linear dependence, respectively.

Specifically, we randomly select a subset of channel pairs \((i, j)\) and compute their MI and PC values based on their gradients \( G_i^{(l)} \) and \( G_j^{(l)} \) at layer \( l \). Instead of fixed thresholds, we define adaptive thresholds \(\tau_{\text{MI}}\) and \(\tau_{\text{corr}}\) as:
\begin{align}
\tau_{\text{MI}} = \beta_{\text{MI}} \cdot \text{mean}(\mathrm{MI}(G_i^{(l)}, G_j^{(l)})), \\
\tau_{\text{corr}} = \beta_{\text{corr}} \cdot \text{mean}(|\mathrm{corr}(G_i^{(l)}, G_j^{(l)})|),
\end{align}
where \(\beta_{\text{MI}}\) and \(\beta_{\text{corr}}\) are scaling factors controlling sensitivity.

If the MI or the absolute PC between a channel pair exceeds its respective threshold, this pair is considered redundant and contributes to suppressing the corresponding channel gradients. The suppression weight \(w_i\) for channel \(i\) in layer \(l\) is computed by aggregating redundancy indicators \(t_{\text{MI}}^{i,j}\) and \(t_{\text{corr}}^{i,j}\) from all sampled pairs \(P\):
\begin{align}
w_i = 1 - \alpha \sum_{(i,j) \in P} (t_{\text{MI}}^{i,j} + t_{\text{corr}}^{i,j}),
\end{align}
where \(\alpha\) is a fixed reduction factor. To prevent excessive suppression, the weight is truncated:
\begin{align}
w_i = \max(0.1, w_i).
\end{align}
Finally, the gradient is adjusted by
\begin{align}
\tilde{G}_i^{(l)} = G_i^{(l)} \cdot w_i.
\end{align}
This adaptive and fine-grained suppression effectively reduces redundant gradient directions, lowering surrogate-specific overfitting and improving adversarial sample transferability.

\paragraph{IS for Additional Knowledge.} 
In data-efficient ViTs, supplementary tokens such as distillation tokens provide extra information that enhances model accuracy but often encode surrogate-specific patterns which harm adversarial transferability.

To mitigate this, we propose an adaptive gradient scaling mechanism for these additional knowledge tokens. Considering the self-attention interaction between primary and additional tokens, the attention can be formulated as
\begin{align}
\text{Attn}(Q, K, V) = \text{softmax}\left(\frac{Q_{\text{primary}} K_{\text{additional}}^\top}{\sqrt{d_k}}\right) V_{\text{additional}},
\end{align}
where \(Q_{\text{primary}}\), \(K_{\text{additional}}\), and \(V_{\text{additional}}\) represent the query, key, and value matrices respectively.

We compute a scaling factor \( c \in (0,1) \) via the sigmoid function on the ratio of L2 norms of their gradients
\begin{equation}
c = \sigma\left(\frac{\|G_{\text{additional}}^{(l)}\|_2}{\|G_{\text{primary}}^{(l)}\|_2}\right),
\end{equation}
which adaptively adjusts the magnitude of gradients from additional tokens:
\begin{equation}
\tilde{G}_{\text{additional}}^{(l)} = c \cdot G_{\text{additional}}^{(l)}.
\end{equation}
This reduces the influence of surrogate-specific information encoded in additional tokens, effectively enhancing the transferability of generated adversarial examples.

\section{Experiments}
\label{sec:experiments}

\subsection{Experiment Setting}

To validate the effectiveness of COGO, we conducted extensive experiments in various models. In this section, we describe the details of our experimental setup, including datasets, models, adversarial attack methods, evaluation metrics, and hyperparameter settings.

\paragraph{Datasets.} Following TGR~\cite{zhang2023transferable} and other prior works, we use 1,000 randomly sampled images from the ILSVRC 2012 validation set~\cite{russakovsky2015imagenet}, a common setting for fair comparison in transfer-based attack studies.

\paragraph{Models.} We evaluate the transferability of adversarial examples generated by different source models in two scenarios: intra-architecture transferability, where ViT-based source and target models are used, and cross-architecture transferability, where ViT-based models are applied to attack CNN target models.

For our experiments, we selected four source models: Visformer-S~\cite{chen2021visformer}, DeiT-B~\cite{touvron2021going}, CaiT-S/24~\cite{touvron2021going} and ViT-B/16~\cite{dosovitskiy2020image}. And the target models include additional ViT architectures, such as TNT-S~\cite{han2021transformer} and ConViT-B~\cite{d2021convit}. For CNN target models, we used both defended and undefended versions. Undefended models include well-known architectures such as Inception-v3~\cite{szegedy2016rethinking}, Inception-v4~\cite{szegedy2017inception}, Inception-ResNet-v2~\cite{szegedy2017inception}, and ResNet-101~\cite{he2016deep}, while the defended models employ adversarial training and other defense techniques~\cite{madry2017towards}, including ensembles of adversarially trained Inception-v3 and Inception-ResNet-v2 models.

\paragraph{Attack Methods.} We evaluate the effectiveness of our COGO by comparing it with several competitive baselines. The primary baseline is TGR~\cite{zhang2023transferable}, which is specifically designed to generate transferable adversarial attacks on ViTs. We also include two state-of-the-art methods, GNS-HFA~\cite{zhuenhancing} and ATT~\cite{ming2024boosting}. In addition, we compare against other strong baselines, including MIM~\cite{dong2018boosting}, SINI-FGSM~\cite{lin2019nesterov}, PNA~\cite{wei2022towards}, and SSA~\cite{long2022frequency}.

\paragraph{Evaluation Metrics.} We measure the attack success rate (ASR) of adversarial examples on different models, which quantifies the proportion of successful attacks that lead to misclassifications by the target model. This is expressed as:
\begin{equation} 
\text{ASR} = \frac{\text{Number of successful attacks}}{\text{Total number of attacks}} \times 100\% 
\end{equation}

\paragraph{Hyperparameter Settings.} In all experiments, hyperparameters were carefully chosen to ensure fairness. Each attack ran for 10 iterations with a maximum \( \ell_\infty \)-norm perturbation of \( \epsilon = 8 \) (scale 0--255). 

\begin{table*}[ht]
    \caption{Performance Comparison of Different Adversarial Attacks on ViT-based and CNN-based Models.}
    \label{tab:adversarial_attack_results}
    \vskip 0.15in
    \small
    \centering
    \resizebox{\textwidth}{!}{
    \begin{tabular}{l|l|cccccc|ccccccc}
        \toprule
        \multicolumn{1}{c|}{\multirow{2}{*}{Model}} & \multicolumn{1}{c|}{\multirow{2}{*}{Attack}} & \multicolumn{6}{c|}{ViT-based Models} & \multicolumn{7}{c}{CNN-based Models} \\
        & & ViT-B/16 & CaiT-S/24 & Visformer-S & DeiT-B & TNT-S & ConViT-B & Inc-v3 & Inc-v4 & IncRes-v2 & Res & \(\text{Inc-v3}_{\text{ens3}}\) & \(\text{Inc-v3}_{\text{ens4}}\) & \(\text{IncRes-v2}_{\text{ens3}}\) \\
        \midrule
        \multirow{8}{*}{Visformer-S}
        & MIM & 18.5 & 24.9 & 99.6 & 20.8 & 41.9 & 23.3 & 31.2 & 31.9 & 22.4 & 25.0 & 11.5 & 10.7 & 6.9 \\
        & SINI-FGSM & 25.8 & 35.6 & 99.7 & 34.1 & 53.7 & 35.9 & 42.3 & 43.0 & 32.7 & 35.4 & 21.0 & 18.1 & 14.5 \\
        & PNA & 19.4 & 28.3 & \textbf{100.0} & 23.6 & 49.9 & 24.4 & 37.4 & 36.0 & 23.7 & 30.9 & 11.3 & 8.9 & 6.2 \\
        & SSA & 22.8 & 28.2 & 96.3 & 27.1 & 55.4 & 26.3 & 43.8 & 44.6 & 33.4 & 32.1 & 21.1 & 16.7 & 12.5 \\
        & TGR & 30.3 & 41.4 & \textbf{100.0} & 36.8 & 66.2 & 35.3 & 49.6 & 54.1 & 34.8 & 50.4 & 22.8 & 18.9 & 11.3 \\
        & GNS-HFA & \underline{49.1} & \underline{58.1} & \underline{99.9} & \underline{54.1} & \underline{81.3} & \underline{52.3} & \underline{71.6} & \underline{71.3} & \underline{57.9} & \underline{68.4} & \underline{48.7} & \underline{47.0} & \underline{34.2} \\
        & ATT & 34.7 & 47.2 & 99.7 & 39.6 & 71.5 & 38.9 & 58.4 & 60.3 & 40.7 & 57.0 & 29.4 & 24.0 & 16.5 \\
        & COGO & \textbf{55.2} & \textbf{66.9} & \textbf{100.0} & \textbf{64.9} & \textbf{85.5} & \textbf{62.0} & \textbf{71.8} & \textbf{72.4} & \textbf{59.1} & \textbf{72.7} & \textbf{50.5} & \textbf{47.9} & \textbf{36.2} \\
        \midrule
        \multirow{8}{*}{DeiT-B}
        & MIM & 65.0 & 90.8 & 51.2 & \textbf{100.0} & 68.9 & 90.8 & 40.2 & 37.6 & 30.1 & 31.5 & 22.1 & 19.4 & 14.6 \\
        & SINI-FGSM & 68.3 & 90.2 & 53.4 & \underline{99.9} & 72.8 & 90.3 & 43.9 & 39.8 & 34.2 & 33.1 & 26.8 & 24.3 & 18.2 \\
        & PNA & 61.3 & 80.8 & 52.1 & 91.1 & 67.0 & 81.0 & 42.1 & 38.8 & 32.0 & 34.0 & 22.2 & 19.6 & 15.0 \\
        & SSA & 68.9 & 92.0 & 44.7 & \underline{99.9} & 75.2 & 90.5 & 47.9 & 45.2 & 40.3 & 30.5 & 25.0 & 20.5 & 16.9 \\
        & TGR & 82.0 & 93.3 & 65.3 & \underline{99.9} & 84.0 & 93.1 & 52.9 & 49.5 & 40.6 & 45.9 & 32.9 & 30.4 & 22.4 \\
        & GNS-HFA & \underline{82.9} & 93.6 & \underline{69.2} & 99.5 & \underline{87.2} & 93.6 & \underline{61.9} & \underline{56.8} & \underline{50.9} & \underline{52.1} & \underline{42.0} & \underline{33.8} & \underline{25.7} \\
        & ATT & 82.5 & \underline{94.7} & \underline{69.2} & 99.8 & 84.9 & \underline{94.9} & 56.3 & 51.5 & 41.6 & 49.8 & 34.1 & 32.2 & 24.5 \\
        & COGO & \textbf{87.6} & \textbf{96.0} & \textbf{73.9} & \underline{99.9} & \textbf{91.3} & \textbf{96.4} & \textbf{64.6} & \textbf{60.1} & \textbf{52.9} & \textbf{56.2} & \textbf{45.3} & \textbf{35.4} & \textbf{27.6} \\
        \midrule
        \multirow{8}{*}{CaiT-S/24}
        & MIM & 48.8 & 99.2 & 34.4 & 69.7 & 54.2 & 68.7 & 33.6 & 29.6 & 22.4 & 21.8 & 15.1 & 12.6 & 9.8 \\
        & SINI-FGSM & 52.2 & 98.3 & 35.0 & 71.5 & 54.7 & 70.3 & 35.9 & 33.3 & 25.8 & 22.3 & 17.8 & 15.8 & 11.8 \\
        & PNA & 44.3 & 82.3 & 33.5 & 61.8 & 52.6 & 60.8 & 32.8 & 29.6 & 22.7 & 22.9 & 14.2 & 11.0 & 8.8 \\
        & SSA & 52.5 & 96.0 & 34.0 & 72.6 & 62.1 & 68.4 & 39.5 & 38.0 & 31.9 & 24.6 & 19.7 & 16.5 & 12.2 \\
        & TGR & 78.9 & \textbf{99.9} & 60.1 & 90.0 & 82.6 & 89.7 & 49.6 & 48.0 & 35.8 & 38.8 & 26.9 & 23.6 & 16.2 \\
        & GNS-HFA & \underline{79.7} & 99.6 & \underline{65.2} & 88.4 & \underline{84.3} & 88.2 & \underline{61.9} & \underline{57.6} & \underline{47.9} & \underline{47.8} & \underline{38.9} & \underline{38.5} & \textbf{28.9} \\
        & ATT & 78.0 & 99.4 & 63.7 & \underline{89.1} & 83.9 & \underline{88.3} & 56.2 & 52.1 & 40.1 & 45.3 & 30.8 & 29.1 & 20.7 \\
        & COGO & \textbf{84.8} & \underline{99.8} & \textbf{68.6} & \textbf{93.0} & \textbf{89.5} & \textbf{92.9} & \textbf{64.0} & \textbf{60.6} & \textbf{49.5} & \textbf{50.0} & \textbf{42.0} & \textbf{39.2} & \underline{28.4} \\
        \midrule
        \multirow{8}{*}{ViT-B/16}
        & MIM & \underline{99.6} & 53.7 & 23.1 & 51.2 & 55.3 & 54.4 & 30.9 & 30.6 & 24.8 & 16.0 & 13.4 & 10.7 & 8.0 \\
        & SINI-FGSM & 98.8 & 56.4 & 27.9 & 55.5 & 61.0 & 56.9 & 36.2 & 35.7 & 29.2 & 16.7 & 15.7 & 13.2 & 9.2 \\
        & PNA & 97.1 & 62.1 & 30.5 & 60.0 & 62.6 & 60.5 & 34.6 & 32.4 & 28.3 & 19.9 & 12.4 & 11.0 & 8.7 \\
        & SSA & 99.4 & 51.8 & 21.5 & 47.5 & 62.0 & 50.5 & 34.6 & 31.4 & 28.6 & 13.5 & 11.1 & 10.0 & 7.5 \\
        & TGR & 99.4 & 56.2 & 30.6 & 54.2 & 66.3 & 57.8 & 37.3 & 36.3 & 29.7 & 19.7 & 16.1 & 14.3 & 10.4 \\
        & GNS-HFA & 98.8 & 65.7 & 38.2 & 64.6 & \underline{73.1} & 67.2 & \underline{44.5} & \underline{41.6} & \underline{37.8} & \underline{25.9} & \underline{21.1} & \underline{19.8} & \underline{15.0} \\
        & ATT & 99.5 & \underline{69.0} & \underline{40.4} & \underline{67.2} & \underline{73.1} & \underline{68.0} & 41.8 & 40.9 & 35.0 & 25.6 & 19.0 & 17.2 & 13.0 \\
        & COGO & \textbf{99.8} & \textbf{81.5} & \textbf{48.2} & \textbf{80.7} & \textbf{80.4} & \textbf{81.2} & \textbf{50.3} & \textbf{50.3} & \textbf{41.8} & \textbf{35.9} & \textbf{27.9} & \textbf{26.1} & \textbf{19.4} \\
        \bottomrule
    \end{tabular}
    }
\end{table*}

\subsection{Results}

In this section, we evaluate the performance of our proposed methods against three categories of models: undefended ViTs, undefended CNNs, and adversarially trained CNNs. Specifically, we generate adversarial examples using a given surrogate model and attack other models in a black-box scenario, as well as the surrogate model itself in a white-box scenario. The results are shown in Table~\ref{tab:adversarial_attack_results}, where the best values in each column are highlighted in bold, and the second-best are underlined.

First, we evaluate transferability across different ViT models. In black-box settings, our strategy achieves an average improvement of 7.2\% in attack success rates over GNS-HFA and an 10.1\% enhancement over the ATT method.

Next, we evaluate the effectiveness of COGO against both undefended and adversarially trained CNNs to assess cross-architecture transferability. Despite the significant architectural differences between ViTs and CNNs, COGO achieves an average improvement of 2.3\% over GNS-HFA and 10.5\% over ATT.

\begin{figure}[ht]
\centering
\includegraphics[width=0.6\linewidth]{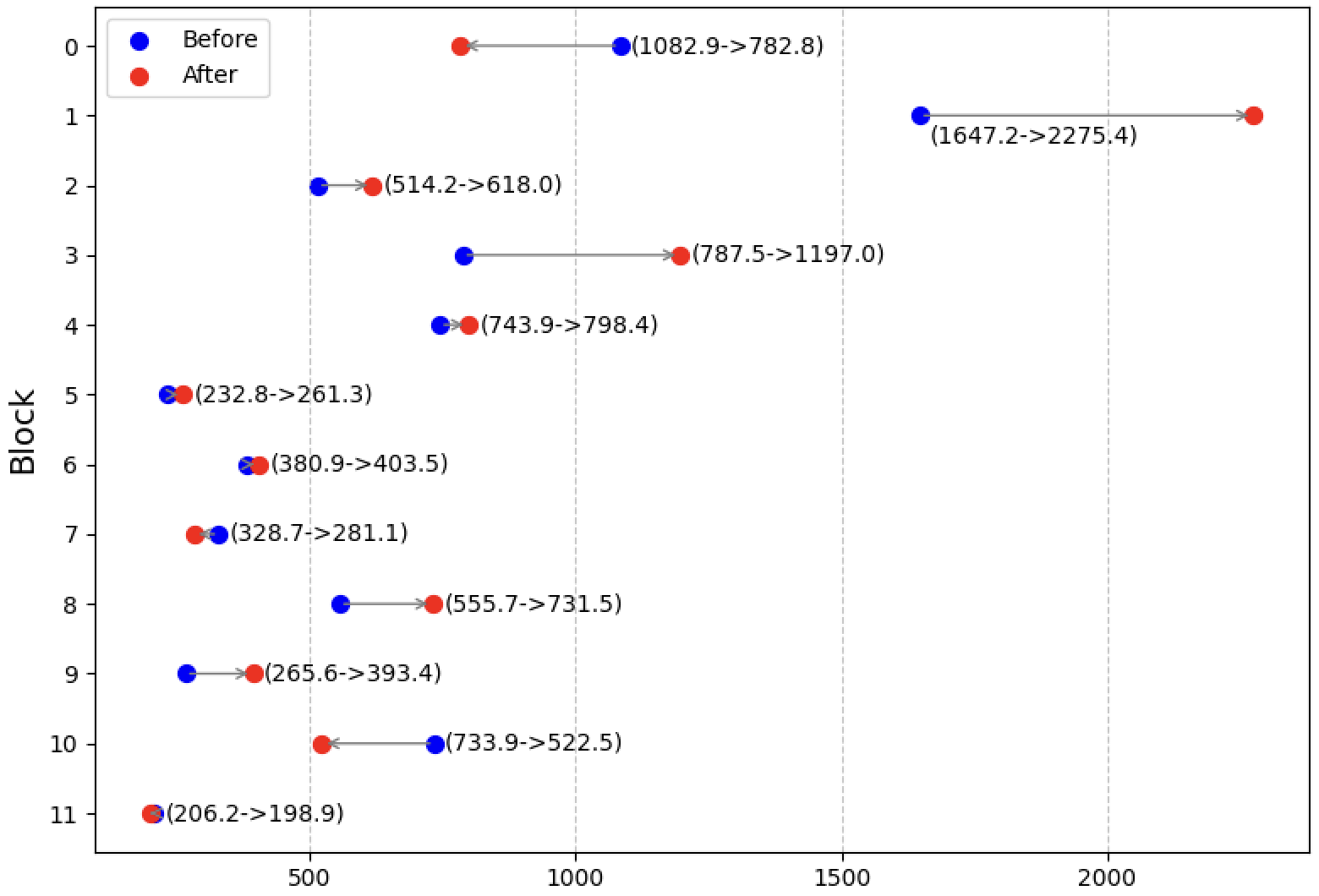}
\caption{Changes in the gradient dispersion indicator across DeiT blocks before and after applying COGO. Larger values reflect more uniform gradient distributions.}
\label{fig:gradient_entropy}
\end{figure}

\begin{table}[ht]
    \caption{Attack success rates (\%) on ViTs, CNNs, and adv-CNNs under different settings of CE and IS. A checkmark (\checkmark) indicates the corresponding component is used, while a dash (-) means it is not applied.
    }
    \label{tab:ablation_fga_he}
    \centering
    \resizebox{0.6\textwidth}{!}{
        \small
        \begin{tabular}{c|c|c|c|c}
        \toprule
        CE & IS & ViTs & CNNs & CNNs-adv \\
        \midrule
        - & - & 46.64 & 30.45 & 9.80 \\
        $\checkmark$ & - & 72.56 (+25.92) & 56.18 (+25.73) & 32.15 (+22.35) \\
        - & $\checkmark$ & 62.38 (+15.74) & 45.85 (+15.40) & 22.77 (+12.97) \\
        $\checkmark$ & $\checkmark$ & 77.97 (+31.33) & 63.73 (+33.28) & 36.75 (+26.95) \\
        \bottomrule
        \end{tabular}
    }
\end{table}

\subsection{Analysis}

To better understand the impact of COGO, we analyzed the gradient flow in ViT models during adversarial example generation. Without COGO, gradients in the \(qkv\) module were highly concentrated on specific tokens and channels tied to dominant surrogate model features, causing perturbations to be overly model-specific and less transferable.

Applying COGO resulted in a more diverse gradient distribution across model blocks. To quantify this, we computed a dispersion-related indicator by flattening and normalizing gradient magnitudes into a probability-like vector and summing its log-weighted elements. This value reflects how broadly the gradients are spread across different dimensions. A larger value indicates a more balanced distribution and reduced dependence on surrogate-specific features. As shown in Figure~\ref{fig:gradient_entropy}, most blocks show increased values after applying COGO, suggesting improved gradient diversity. A few blocks, such as \texttt{blocks[0]} and \texttt{blocks[10]}, show slightly reduced values due to the CE component, which enhances shared mid-to-low frequency features and causes gradient concentration in those regions. This complementarity between IS and CE promotes more transferable and model-common features.

Overall, the more balanced and diverse gradient landscape induced by COGO helps generate adversarial perturbations that better capture transferable features, thus improving attack success rates across different models.

\begin{table*}[ht]
    \caption{Performance Comparison using Different Enhancement Coefficients $\gamma$ on ViT-B/16.}
    \label{tab:compare_gamma}
    \small
    \vskip 0.15in
    \centering
    \resizebox{\textwidth}{!}{
    \begin{tabular}{l|cccccc|ccccccc}
        \toprule
        \multicolumn{1}{c|}{\multirow{2}{*}{$\gamma$}} & \multicolumn{6}{c|}{ViT-based Models} & \multicolumn{7}{c}{CNN-based Models} \\
        & ViT-B/16 & CaiT-S/24 & Visformer-S & DeiT-B & TNT-S & ConViT-B & Inc-v3 & Inc-v4 & IncRes-v2 & Res & \(\text{Inc-v3}_{\text{ens3}}\) & \(\text{Inc-v3}_{\text{ens4}}\) & \(\text{IncRes-v2}_{\text{ens3}}\) \\
        \midrule
        -0.25 & 100.0 & 72.5 & 41.1 & 73.3 & 76.6 & 73.7 & 49.0 & 44.7 & 41.7 & 27.7 & 23.2 & 20.6 & 14.8 \\
        -0.1 & 100.0 & 78.9 & 46.5 & 78.3 & 78.3 & 78.7 & \textbf{52.2} & 47.4 & \textbf{42.6} & 32.6 & 25.4 & 21.8 & 17.1 \\
        0 & 98.8 & 65.7 & 38.2 & 64.6 & 73.1 & 67.2 & 44.5 & 41.6 & 37.8 & 25.9 & 21.1 & 19.8 & 15.0 \\
        0.1 & 99.6 & 71.9 & 43.5 & 70.9 & 76.9 & 73.7 & 49.7 & 45.9 & 39.6 & 29.3 & 22.8 & 21.1 & 16.7 \\
        0.25 & 99.6 & 74.6 & 45.6 & 73.9 & 79.0 & 76.4 & 50.4 & 48.1 & 41.0 & 31.3 & 25.1 & 22.9 & 18.7 \\
        0.5 & 99.5 & 77.4 & 46.1 & 76.2 & 79.9 & 78.4 & 50.2 & 48.0 & 40.6 & 33.7 & 26.2 & 24.8 & 18.2 \\
        0.75 & 99.4 & 74.8 & 45.8 & 74.5 & 79.4 & 75.3 & 48.6 & 47.3 & 39.7 & 32.2 & 26.0 & 23.1 & 18.0 \\
        1 & 99.8 & \textbf{81.5} & \textbf{48.2} & \textbf{80.7} & \textbf{80.4} & \textbf{81.2} & 50.3 & \textbf{50.3} & 41.8 & \textbf{35.9} & \textbf{27.9} & \textbf{26.1} & \textbf{19.4} \\
        1.5 & 97.9 & 67.6 & 38.0 & 67.6 & 73.2 & 65.7 & 44.6 & 44.5 & 37.9 & 28.4 & 22.7 & 21.0 & 15.7 \\
        \bottomrule
    \end{tabular}}
\end{table*}

\begin{figure*}[ht]
    \centering
    \subfloat[Original]{\includegraphics[width=0.18\textwidth]{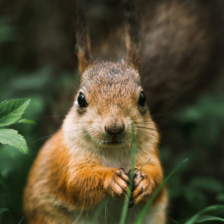}}    
    \hfill
    \subfloat[N=2]{\includegraphics[width=0.18\textwidth]{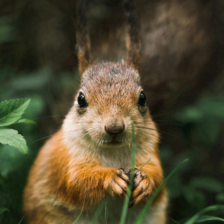}}
    \hfill
    \subfloat[N=6]{\includegraphics[width=0.18\textwidth]{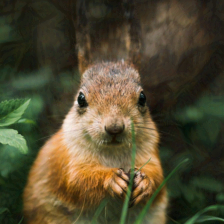}}
    \hfill
    \subfloat[N=10]{\includegraphics[width=0.18\textwidth]{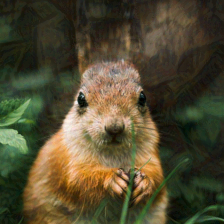}}
    \hfill
    \subfloat[N=14]{\includegraphics[width=0.18\textwidth]{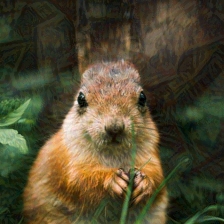}}

    \caption{Adversarial attack examples at different N}
    \label{fig:impact_N}
\end{figure*}

\subsection{Ablation Study}

\paragraph{Impact of CE and IS.}  
Ablation results in Table~\ref{tab:ablation_fga_he} demonstrate that both CE and IS significantly improve attack success rates on ViTs, CNNs, and adversarially trained CNNs. Among them, CE has a more pronounced effect. Combining CE and IS achieves the highest success rates, highlighting their complementary roles. This indicates that guiding perturbations deliberately is most effective when the main challenges are clearly addressed.

\paragraph{Impact of enhancement coefficient \(\gamma\) in CE.}  
The parameter \(\gamma\) controls the strength of perturbations applied to mid-to-low frequency DCT coefficients (Eq.~\ref{eq:dct}). Table~\ref{tab:compare_gamma} shows that \(\gamma=1\) yields the best balance between perturbation strength and transferability, outperforming both higher (\(\gamma=1.5\)) and lower (\(\gamma=0.1\)) values, which either cause over-perturbation or insufficient enhancement.

\paragraph{Impact of iteration number \(N\).}  
As illustrated in Figure~\ref{fig:impact_N}, increasing the number of iterations \(N\) leads to more complex perturbations. However, improvements plateau around \(N=10\). Therefore, to balance computational cost and attack effectiveness, we choose \(N=10\) as the default setting.

\paragraph{Impact of other hyperparameters.}  
We also examined other hyperparameters such as the number of selected channels, reduction thresholds for MI and PC, and the step size constant. Detailed results and analyses are provided in Appendix~\ref{sec:other_ablation}.

\section{Conclusion}
\label{sec:conc}

In this paper, we propose \textbf{\textit{COGO}}, a commonality-oriented gradient optimization strategy to enhance adversarial attack transferability on ViTs. Our method improves transferability by boosting perturbations related to features shared across surrogate models via frequency-domain noise injection (CE) and suppressing gradients linked to surrogate-specific characteristics using adaptive thresholding (IS). Extensive experiments on ViTs and CNNs demonstrate that \textbf{\textit{COGO}} significantly outperforms existing approaches.

\bibliographystyle{splncs04}
\bibliography{LaTeX2e_Proceedings_Templates/main}

\begin{thebibliography}{10}
\providecommand{\url}[1]{\texttt{#1}}
\providecommand{\urlprefix}{URL }
\providecommand{\doi}[1]{https://doi.org/#1}

\bibitem{chen2021visformer}
Chen, Z., Xie, L., Niu, J., Liu, X., Wei, L., Tian, Q.: Visformer: The vision-friendly transformer. In: Proceedings of the IEEE/CVF international conference on computer vision. pp. 589--598 (2021)

\bibitem{dong2018boosting}
Dong, Y., Liao, F., Pang, T., Su, H., Zhu, J., Hu, X., Li, J.: Boosting adversarial attacks with momentum. In: Proceedings of the IEEE conference on computer vision and pattern recognition. pp. 9185--9193 (2018)

\bibitem{dong2019evading}
Dong, Y., Pang, T., Su, H., Zhu, J.: Evading defenses to transferable adversarial examples by translation-invariant attacks. In: Proceedings of the IEEE/CVF conference on computer vision and pattern recognition. pp. 4312--4321 (2019)

\bibitem{dosovitskiy2020image}
Dosovitskiy, A., Beyer, L., Kolesnikov, A., Weissenborn, D., Zhai, X., Unterthiner, T., Dehghani, M., Minderer, M., Heigold, G., Gelly, S., et~al.: An image is worth 16x16 words: Transformers for image recognition at scale. arXiv preprint arXiv:2010.11929  (2020)

\bibitem{d2021convit}
d’Ascoli, S., Touvron, H., Leavitt, M.L., Morcos, A.S., Biroli, G., Sagun, L.: Convit: Improving vision transformers with soft convolutional inductive biases. In: International conference on machine learning. pp. 2286--2296. PMLR (2021)

\bibitem{fu2024fast}
Fu, K., Qu, L., Wang, S., Xiong, Y., Maglogiannis, I., Gao, L., Wang, M., et~al.: Fast: A dual-tier few-shot learning paradigm for whole slide image classification. Advances in Neural Information Processing Systems  \textbf{37},  105090--105113 (2024)

\bibitem{fu2025dual}
Fu, K., Yuan, M., Wang, C., Pang, W., Chi, J., Wang, M., Gao, L.: Dual focus-attention transformer for robust point cloud registration. In: Proceedings of the Computer Vision and Pattern Recognition Conference. pp. 11769--11778 (2025)

\bibitem{goodfellow2014explaining}
Goodfellow, I.J.: Explaining and harnessing adversarial examples. arXiv preprint arXiv:1412.6572  (2014)

\bibitem{hamdi2020advpc}
Hamdi, A., Rojas, S., Thabet, A., Ghanem, B.: Advpc: Transferable adversarial perturbations on 3d point clouds. In: Computer Vision--ECCV 2020: 16th European Conference, Glasgow, UK, August 23--28, 2020, Proceedings, Part XII 16. pp. 241--257. Springer (2020)

\bibitem{han2021transformer}
Han, K., Xiao, A., Wu, E., Guo, J., Xu, C., Wang, Y.: Transformer in transformer. Advances in neural information processing systems  \textbf{34},  15908--15919 (2021)

\bibitem{he2024theoretically}
He, H., Liu, Y., Wang, Z., Mao, Y., Bu, Y.: Theoretically grounded framework for llm watermarking: A distribution-adaptive approach. arXiv preprint arXiv:2410.02890  (2024)

\bibitem{he2016deep}
He, K., Zhang, X., Ren, S., Sun, J.: Deep residual learning for image recognition. In: Proceedings of the IEEE conference on computer vision and pattern recognition. pp. 770--778 (2016)

\bibitem{heo2021rethinking}
Heo, B., Yun, S., Han, D., Chun, S., Choe, J., Oh, S.J.: Rethinking spatial dimensions of vision transformers. In: Proceedings of the IEEE/CVF international conference on computer vision. pp. 11936--11945 (2021)

\bibitem{inkawhich2020transferable}
Inkawhich, N., Liang, K.J., Carin, L., Chen, Y.: Transferable perturbations of deep feature distributions. arXiv preprint arXiv:2004.12519  (2020)

\bibitem{kim2024exploring}
Kim, G., Kim, J., Lee, J.S.: Exploring adversarial robustness of vision transformers in the spectral perspective. In: Proceedings of the IEEE/CVF Winter Conference on Applications of Computer Vision. pp. 3976--3985 (2024)

\bibitem{kurakin2018adversarial}
Kurakin, A., Goodfellow, I.J., Bengio, S.: Adversarial examples in the physical world. In: Artificial intelligence safety and security, pp. 99--112. Chapman and Hall/CRC (2018)

\bibitem{lin2019nesterov}
Lin, J., Song, C., He, K., Wang, L., Hopcroft, J.E.: Nesterov accelerated gradient and scale invariance for adversarial attacks. arXiv preprint arXiv:1908.06281  (2019)

\bibitem{lin2023advances}
Lin, J., Gao, H., Feng, X., Xu, R., Wang, C., Zhang, M., Guo, L., Xu, S.: Advances in embodied navigation using large language models: A survey. arXiv preprint arXiv:2311.00530  (2023)

\bibitem{Liu_2025_VaLiK}
Liu, J., Meng, S., Gao, Y., Mao, S., Cai, P., Yan, G., Chen, Y., Bian, Z., Shi, B., Wang, D.: Aligning vision to language: Text-free multimodal knowledge graph construction for enhanced llms reasoning. arXiv preprint arXiv:2503.12972  (2025)

\bibitem{liu2016delving}
Liu, Y., Chen, X., Liu, C., Song, D.: Delving into transferable adversarial examples and black-box attacks. arXiv preprint arXiv:1611.02770  (2016)

\bibitem{liu2024adaptive}
Liu, Y., Bu, Y.: Adaptive text watermark for large language models. arXiv preprint arXiv:2401.13927  (2024)

\bibitem{liu2024image}
Liu, Y., Song, Y., Ci, H., Zhang, Y., Wang, H., Shou, M.Z., Bu, Y.: Image watermarks are removable using controllable regeneration from clean noise. arXiv preprint arXiv:2410.05470  (2024)

\bibitem{liu2025dataset}
Liu, Y., Zhao, X., Song, D., Bu, Y.: Dataset protection via watermarked canaries in retrieval-augmented llms. arXiv preprint arXiv:2502.10673  (2025)

\bibitem{liu2021swin}
Liu, Z., Lin, Y., Cao, Y., Hu, H., Wei, Y., Zhang, Z., Lin, S., Guo, B.: Swin transformer: Hierarchical vision transformer using shifted windows. In: Proceedings of the IEEE/CVF international conference on computer vision. pp. 10012--10022 (2021)

\bibitem{long2022frequency}
Long, Y., Zhang, Q., Zeng, B., Gao, L., Liu, X., Zhang, J., Song, J.: Frequency domain model augmentation for adversarial attack. In: European conference on computer vision. pp. 549--566. Springer (2022)

\bibitem{lou2023trojtext}
Lou, Q., Liu, Y., Feng, B.: Trojtext: Test-time invisible textual trojan insertion. arXiv preprint arXiv:2303.02242  (2023)

\bibitem{madry2017towards}
Madry, A.: Towards deep learning models resistant to adversarial attacks. arXiv preprint arXiv:1706.06083  (2017)

\bibitem{ming2024boosting}
Ming, D., Ren, P., Wang, Y., Feng, X.: Boosting the transferability of adversarial attack on vision transformer with adaptive token tuning. Advances in Neural Information Processing Systems  \textbf{37},  20887--20918 (2024)

\bibitem{moosavi2017universal}
Moosavi-Dezfooli, S.M., Fawzi, A., Fawzi, O., Frossard, P.: Universal adversarial perturbations. In: Proceedings of the IEEE conference on computer vision and pattern recognition. pp. 1765--1773 (2017)

\bibitem{papernot2017practical}
Papernot, N., McDaniel, P., Goodfellow, I., Jha, S., Celik, Z.B., Swami, A.: Practical black-box attacks against machine learning. In: Proceedings of the 2017 ACM on Asia conference on computer and communications security. pp. 506--519 (2017)

\bibitem{raghu2021vision}
Raghu, M., Unterthiner, T., Kornblith, S., Zhang, C., Dosovitskiy, A.: Do vision transformers see like convolutional neural networks? Advances in neural information processing systems  \textbf{34},  12116--12128 (2021)

\bibitem{russakovsky2015imagenet}
Russakovsky, O., Deng, J., Su, H., Krause, J., Satheesh, S., Ma, S., Huang, Z., Karpathy, A., Khosla, A., Bernstein, M., et~al.: Imagenet large scale visual recognition challenge. International journal of computer vision  \textbf{115},  211--252 (2015)

\bibitem{shao2021adversarial}
Shao, R., Shi, Z., Yi, J., Chen, P.Y., Hsieh, C.J.: On the adversarial robustness of vision transformers. arXiv preprint arXiv:2103.15670  (2021)

\bibitem{shen2025efficient}
Shen, H., Zhang, J., Xiong, B., Hu, R., Chen, S., Wan, Z., Wang, X., Zhang, Y., Gong, Z., Bao, G., et~al.: Efficient diffusion models: A survey. Transactions on Machine Learning Research (TMLR)  (2025)

\bibitem{szegedy2017inception}
Szegedy, C., Ioffe, S., Vanhoucke, V., Alemi, A.: Inception-v4, inception-resnet and the impact of residual connections on learning. In: Proceedings of the AAAI conference on artificial intelligence. vol.~31 (2017)

\bibitem{szegedy2016rethinking}
Szegedy, C., Vanhoucke, V., Ioffe, S., Shlens, J., Wojna, Z.: Rethinking the inception architecture for computer vision. In: Proceedings of the IEEE conference on computer vision and pattern recognition. pp. 2818--2826 (2016)

\bibitem{touvron2021training}
Touvron, H., Cord, M., Douze, M., Massa, F., Sablayrolles, A., J{\'e}gou, H.: Training data-efficient image transformers \& distillation through attention. In: International conference on machine learning. pp. 10347--10357. PMLR (2021)

\bibitem{touvron2021going}
Touvron, H., Cord, M., Sablayrolles, A., Synnaeve, G., J{\'e}gou, H.: Going deeper with image transformers. In: Proceedings of the IEEE/CVF international conference on computer vision. pp. 32--42 (2021)

\bibitem{tramer2017ensemble}
Tram{\`e}r, F., Kurakin, A., Papernot, N., Goodfellow, I., Boneh, D., McDaniel, P.: Ensemble adversarial training: Attacks and defenses. arXiv preprint arXiv:1705.07204  (2017)

\bibitem{wan2025srpo}
Wan, Z., Dou, Z., Liu, C., Zhang, Y., Cui, D., Zhao, Q., Shen, H., Xiong, J., Xin, Y., Jiang, Y., et~al.: Srpo: Enhancing multimodal llm reasoning via reflection-aware reinforcement learning. arXiv preprint arXiv:2506.01713  (2025)

\bibitem{wan2025d2o}
Wan, Z., Wu, X., Zhang, Y., Xin, Y., Tao, C., Zhu, Z., Wang, X., Luo, S., Xiong, J., Wang, L., et~al.: D2o: Dynamic discriminative operations for efficient long-context inference of large language models. In: ICLR (2025)

\bibitem{wang2017adversarial}
Wang, B., Yang, Y., Xu, X., Hanjalic, A., Shen, H.T.: Adversarial cross-modal retrieval. In: Proceedings of the 25th ACM international conference on Multimedia. pp. 154--162 (2017)

\bibitem{wang2023accurate}
Wang, C., Xu, R., Xu, S., Meng, W., Xiao, J., Zhang, X.: Accurate lung nodule segmentation with detailed representation transfer and soft mask supervision. IEEE transactions on neural networks and learning systems  (2023)

\bibitem{wei2022towards}
Wei, Z., Chen, J., Goldblum, M., Wu, Z., Goldstein, T., Jiang, Y.G.: Towards transferable adversarial attacks on vision transformers. In: Proceedings of the AAAI Conference on Artificial Intelligence. vol.~36, pp. 2668--2676 (2022)

\bibitem{wu2020visual}
Wu, B., Xu, C., Dai, X., Wan, A., Zhang, P., Yan, Z., Tomizuka, M., Gonzalez, J., Keutzer, K., Vajda, P.: Visual transformers: Token-based image representation and processing for computer vision. arXiv preprint arXiv:2006.03677  (2020)

\bibitem{wu2020skip}
Wu, D., Wang, Y., Xia, S.T., Bailey, J., Ma, X.: Skip connections matter: On the transferability of adversarial examples generated with resnets. arXiv preprint arXiv:2002.05990  (2020)

\bibitem{wu2025segment}
Wu, J., Xu, R., Wood-Doughty, Z., Wang, C., Xu, S., Lam, E.Y.: Segment anything model is a good teacher for local feature learning. IEEE Transactions on Image Processing  (2025)

\bibitem{xie2019improving}
Xie, C., Zhang, Z., Zhou, Y., Bai, S., Wang, J., Ren, Z., Yuille, A.L.: Improving transferability of adversarial examples with input diversity. In: Proceedings of the IEEE/CVF conference on computer vision and pattern recognition. pp. 2730--2739 (2019)

\bibitem{xu2024skinformer}
Xu, R., Wang, C., Zhang, J., Xu, S., Meng, W., Zhang, X.: Skinformer: Learning statistical texture representation with transformer for skin lesion segmentation. IEEE Journal of Biomedical and Health Informatics  \textbf{28}(10),  6008--6018 (2024)

\bibitem{xu2024mrftrans}
Xu, R., Zhang, J., Sun, J., Wang, C., Wu, Y., Xu, S., Meng, W., Zhang, X.: Mrftrans: Multimodal representation fusion transformer for monocular 3d semantic scene completion. Information Fusion  \textbf{111},  102493 (2024)

\bibitem{xue2023trojllm}
Xue, J., Zheng, M., Hua, T., Shen, Y., Liu, Y., B{\"o}l{\"o}ni, L., Lou, Q.: Trojllm: A black-box trojan prompt attack on large language models. Advances in Neural Information Processing Systems  \textbf{36},  65665--65677 (2023)

\bibitem{zhang2023transferable}
Zhang, J., Huang, Y., Wu, W., Lyu, M.R.: Transferable adversarial attacks on vision transformers with token gradient regularization. In: Proceedings of the IEEE/CVF Conference on Computer Vision and Pattern Recognition. pp. 16415--16424 (2023)

\bibitem{zhang2024mg}
Zhang, Y., Liu, Y., Miao, D., Zhang, Q., Shi, Y., Hu, L.: Mg-vit: a multi-granularity method for compact and efficient vision transformers. Advances in Neural Information Processing Systems  \textbf{36} (2024)

\bibitem{zhang2024mlip}
Zhang, Y., Zhang, Q., Gong, Z., Shi, Y., Liu, Y., Miao, D., Liu, Y., Liu, K., Yi, K., Fan, W., et~al.: Mlip: Efficient multi-perspective language-image pretraining with exhaustive data utilization. arXiv preprint arXiv:2406.01460  (2024)

\bibitem{zhang2025enhancing}
Zhang, Y., Zhou, J., Li, X., Zhang, Q., Wan, Z., Wang, T., Miao, D., Wang, C., Cao, L.: Enhancing text-to-image diffusion transformer via split-text conditioning. arXiv preprint arXiv:2505.19261  (2025)

\bibitem{zhang2025cae}
Zhang, Z., Wang, C., Xu, R., Xu, W., Xu, S., Zhang, Y., Guo, L.: Cae-dfkd: Bridging the transferability gap in data-free knowledge distillation. arXiv preprint arXiv:2504.21478  (2025)

\bibitem{zhuenhancing}
Zhu, Z., Wang, X., Jin, Z., Zhang, J., Chen, H.: Enhancing transferable adversarial attacks on vision transformers through gradient normalization scaling and high-frequency adaptation. In: The Twelfth International Conference on Learning Representations (2024)

\end{thebibliography}

\clearpage
\setcounter{page}{1}

\renewcommand{\thesection}{\Alph{section}}
\setcounter{section}{0}

\section{Architectural Module Analysis}
\label{sec:details}

In this section, we provide a detailed analysis of the key modules within ViTs that are potential targets for perturbation application. The analysis is organized in the order that data flows through these modules, examining each module's role and the impact of gradient adjustments systematically.

\paragraph{\(qkv\) Layers} Query \((Q)\), Key \((K)\), and Value \((V)\) layers are essential components of the attention mechanism. They determine how tokens interact with one another to form the Attention Map. Adjusting \(Q\) gradients alters token-to-token query relationships, which can redistribute focus across tokens. Adjusting \(K\) gradients modifies the Attention Map distributions, influencing how features are weighted globally. Adjusting \(V\) gradients directly affects the token output representations, which propagate through the network as features.

\paragraph{Attention Projection Layer} The Attention Projection Layer is responsible for linearly projecting the output of the multi-head attention mechanism back to the token feature space. It consolidates the information processed by all attention heads. Adjusting gradients in this module allows fine-tuning of the final feature distribution derived from the Attention Map, ensuring that perturbations affect a wider range of features rather than over-relying on specific heads or tokens. This makes it a critical target for balancing attention diversity and improving perturbation generalization.

\paragraph{Attention Dropout} Applied to the Attention Map, dropout reduces the dependency on specific token-to-token relationships. Manipulating gradients here disperses the token importance distribution, increasing the diversity of features targeted by the perturbation.

\paragraph{\(mlp\) Layers} Multi-Layer Perceptrons process each token independently, providing non-linear transformations to enrich token features. Gradient adjustments in \(mlp\) layers influence token-level transformations, complementing the global adjustments made in the Attention mechanism.

\begin{table}[ht]
    \caption{Attack success rates for different module combinations on ViTs, CNNs, and adversarially trained CNNs.}
    \label{tab:modules}
    \vskip 0.15in
    \small
    \centering
    \begin{tabular}{l|cccc}
        \toprule
        \multicolumn{1}{c|}{modules} & ViTs & CNNs & CNNs-adv & Average \\
        \midrule
        \(qkv\)  & 66.90 & \textbf{69.00} & \textbf{37.90} & \textbf{57.93} \\
        \(qkv\)+\(proj\)  & \textbf{66.92} & 68.98 & 37.86 & 57.92 \\
        \(qkv\)+\(dropout\) & 57.54 & 57.10 & 28.82 & 47.82 \\
        \(qkv\)+\(mlp\)  & 50.82 & 59.79 & 28.50 & 46.37 \\
        \bottomrule
    \end{tabular}
\end{table}

\vspace{\baselineskip} From the analysis above, it is evident that some modules play a more critical role in determining the effectiveness and transferability of generated perturbations. Among these, the \(qkv\) layers in the Multi-Head Attention mechanism are the most impactful, as they directly influence the Attention Map by controlling token-to-token interactions and feature representations. Extreme gradients or redundant features arising in these layers are more likely to affect the generated perturbations, embedding characteristics that are overly specific to the substitute model.

To further explore the impact of gradient adjustments in other modules, we designed three combinations: \(qkv\) layers with the Attention Projection Layer, \(qkv\) layers with Attention Dropout, and \(qkv\) layers with the \(mlp\) layers. For each combination, we applied our gradient adjustment method and compared the results to the baseline method in our paper, where adjustments were limited to the \(qkv\) layers. These experiments illustrated as Table~\ref{tab:modules}  allow us to evaluate the collaborative effects of gradient adjustments across multiple modules and their contributions to the overall perturbation transferability.

\section{Algorithm}
\label{alg:fga}
The IS algorithm has been demonstrated in the Algorithm~\ref{algorithm:1}.

\begin{algorithm}[]
\caption{Individuality Suppression}
\begin{algorithmic}[1]
\State \textbf{Input:} gradient set $G$, reduction factor $\alpha$, MI threshold scaling factors $\beta_{\text{MI}}$, $\beta_{\text{corr}}$, scaling factor $c$
\State \textbf{Output:} adjusted gradient set $\tilde{G}$
\State $G = [g_1, g_2, \cdots, g_C]$ \Comment{Gradients for redundant features}
\State $G_{\text{additional}} = [g_{\text{additional}}]$ \Comment{Gradients for additional knowledge features}

\State \textbf{Step 1: Suppress Redundant Features}
\State $P$ = Random subset of channel pairs $(i, j)$
\For{$i \in \{1, 2, \cdots, C\}$}
    \For{each $(i, j) \in P$}
        \State Compute $\mathrm{MI}(g_i, g_j)$ and $\mathrm{corr}(g_i, g_j)$
        \State \textbf{if} $\mathrm{MI}(g_i, g_j) > \tau_{\text{MI}}$ \textbf{then} 
            \State \hspace{2em} $t_{\text{MI}}^{i,j} = 1$
        \State \textbf{else} 
            \State \hspace{2em} $t_{\text{MI}}^{i,j} = 0$
        \State \textbf{end if}
        \State \textbf{if} $|\mathrm{corr}(g_i, g_j)| > \tau_{\text{corr}}$ \textbf{then} 
            \State \hspace{2em} $t_{\text{corr}}^{i,j} = 1$
        \State \textbf{else} 
            \State \hspace{2em} $t_{\text{corr}}^{i,j} = 0$
        \State \textbf{end if}
    \EndFor
    \State $s_i = \alpha \sum_{(i,j) \in P} (t_{\text{MI}}^{i,j} + t_{\text{corr}}^{i,j})$
    \State $w_i = \max(1 - s_i, 0.1)$ \Comment{Truncate weight to prevent too small values}
    \State $g_i = g_i \cdot w_i$ \Comment{Adjust gradient for channel $i$}
\EndFor

\State \textbf{Step 2: Suppress Additional Knowledge Features}
\State Compute scaling factor $c = \sigma\left(\frac{\|G_{\text{additional}}\|_2}{\|G_{\text{primary}}\|_2}\right)$
\State $G_{\text{additional}} = G_{\text{additional}} \cdot c$ \Comment{Adjust gradient for additional knowledge features}

\State \textbf{return} $\tilde{G} = [G, G_{\text{additional}}]$ \Comment{Return adjusted gradients}
\end{algorithmic}
\label{algorithm:1}
\end{algorithm}

\section{Hyperparameter Ablation Study}
\label{sec:other_ablation}

\subsection{Impact of the Number of Selected Channel Pairs}
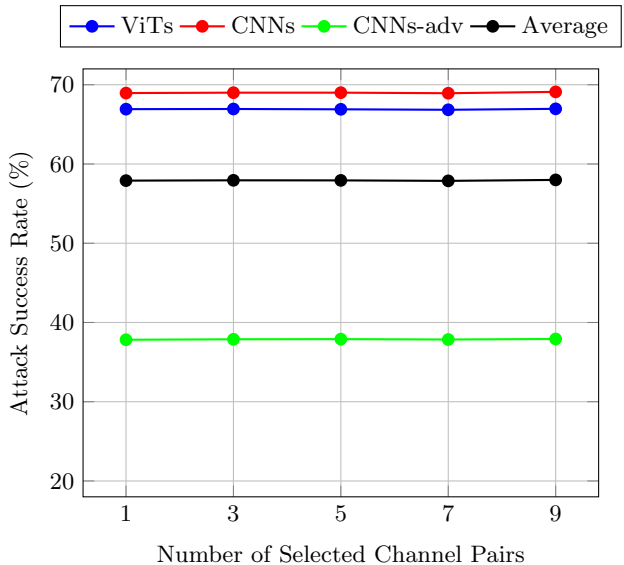
\begin{figure}[ht]
    \centering
    \begin{tikzpicture}
    \begin{axis}[
        xlabel={Number of Selected Channel Pairs},
        ylabel={Attack Success Rate (\%)},
        xtick={1, 3, 5, 7, 9},
        grid=both,
        legend style={font=\small, at={(0.5,1.05)}, anchor=south, legend columns=-1}, 
        ymin=18, ymax=72,
        ytick={20, 30, 40, 50, 60, 70},
        ylabel near ticks
    ]
    
    \addplot[
        color=blue,
        mark=*,
        thick
    ] coordinates {
        (1, 66.92)
        (3, 66.94)
        (5, 66.90)
        (7, 66.84)
        (9, 66.96)
    };
    \addlegendentry{ViTs}
    
    \addplot[
        color=red,
        mark=*,
        thick
    ] coordinates {
        (1, 68.95)
        (3, 69.00)
        (5, 69.00)
        (7, 68.93)
        (9, 69.10)
    };
    \addlegendentry{CNNs}
    
    \addplot[
        color=green,
        mark=*,
        thick
    ] coordinates {
        (1, 37.82)
        (3, 37.88)
        (5, 37.90)
        (7, 37.85)
        (9, 37.92)
    };
    \addlegendentry{CNNs-adv}

    \addplot[
        color=black,
        mark=*,
        thick
    ] coordinates {
        (1, 57.90)
        (3, 57.94)
        (5, 57.93)
        (7, 57.87)
        (9, 57.99)
    };
    \addlegendentry{Average}
    \end{axis}
    \end{tikzpicture}
    \caption{Attack success rates (\%) for different number of selected channel pairs on ViTs, CNNs, adversarially trained CNNs and their average.}
    \label{fig:pairs_comparison}
\end{figure}

To investigate the effect of the number of selected channel pairs ($n$) on gradient adjustments, we carried out experiments by varying $n$ for mutual information and correlation calculations. The default setting in our paper uses 5 channel pairs. However, we first tested additional settings with $n = 1, 3, 7, 9$. The results, summerized in Table~\ref{tab:pairs_comparison} and visualized in Figure~\ref{fig:pairs_comparison}, show that the attack success rates remain largely consistent across different values of $n$. While minor fluctuations are observed, increasing the number of selected channel pairs does not significantly improve the attack success rates.

To ensure this consistency was not due to selecting too few channel pairs, we further conducted an experiment with a significantly larger value, $n = 50$, and summarized the results in Table~\ref{tab:50pairs}.Interestingly, the results for $n = 50$ are comparable to, or even slightly worse than, the results for $n = 9$, suggesting that selecting a larger number of channel pairs does not provide additional benefits. Furthermore, even with $n = 1$, where only a single randomly selected channel pair is used for adjustments, the attack success rates are nearly on par with those achieved using larger values of $n$.

\begin{table}[ht]
    \caption{Attack success rates (\%) for different numbers of selected channel pairs on ViTs, CNNs, and adversarially trained CNNs.}
    \label{tab:pairs_comparison}
    \vskip 0.15in
    \small
    \centering
    \begin{tabular}{c|cccc}
        \toprule
        $n$ & ViTs & CNNs & CNNs-adv & Average \\
        \midrule
        1  & 66.92 & 68.95 & 37.82 & 57.90 \\
        3  & 66.94 & 69.00 & 37.88 & 57.94 \\
        5  & 66.90 & 69.00 & 37.90 & 57.93 \\
        7  & 66.84 & 68.93 & 37.85 & 57.87 \\
        9  & \textbf{66.96} & \textbf{69.10} & \textbf{37.92} & \textbf{57.99} \\
        \bottomrule
    \end{tabular}
\end{table}

\begin{table}[ht]
    \caption{Attack success rates (\%) for selecting 50 channel pairs on ViTs, CNNs, and adversarially trained CNNs.}
    \label{tab:50pairs}
    \vskip 0.15in
    \small
    \centering
    \begin{tabular}{c|cccc}
        \toprule
        $n$ & ViTs & CNNs & CNNs-adv & Average \\
        \midrule
        50  & 66.90 & 69.05 & 37.92 & 57.96 \\
        \bottomrule
    \end{tabular}
\end{table}

\begin{table}[ht]
    \caption{Attack success rates(\%) comparison with different threshold values for MI and PC used in the gradient adjustment function. }
    \label{tab:ablation_hyperparams}
    \vskip 0.15in
    \small
    \centering
    \begin{tabular}{cc|ccc}
        \toprule
        $\tau_{\text{MI}}$ & $\tau_{\text{corr}}$ & \textbf{ViTs} & \textbf{CNNs} & \textbf{CNNs-adv} \\
        \midrule
        0.3 & 0.5 & 66.7 & 68.5 & 37.6 \\
        0.5 & 0.7 & \textbf{66.9} & \textbf{69.0} & \textbf{37.9} \\
        0.7 & 0.9 & 66.2 & 68.4 & 37.2 \\
        \bottomrule
    \end{tabular}
\end{table}

This phenomenon can be attributed to the nature of high-dimensional feature spaces, where most critical gradient information is often concentrated in a small subset of channels. Channels with higher mutual information or strong correlations tend to dominate the gradient flow, as they represent key relationships in the feature representation learned by the model. By adjusting even a small number of these channel pairs, it becomes possible to effectively disrupt redundant and overrepresented features in the gradient signal, thereby optimizing the perturbation's impact.

\begin{table}[ht]
    \caption{Attack success rates (\%) for different scaling factors $c$.}
    \label{tab:distill_token_scaling}
    \vskip 0.15in
    \small
    \centering
    \begin{tabular}{c|cccc}
        \toprule
        $c$ & ViTs & CNNs & CNNs-adv & Average \\
        \midrule
        0.0  & 88.84 & 57.85 & 35.52 & 60.74 \\
        0.1  & \textbf{89.04} & \textbf{58.45} & \textbf{36.10} & \textbf{61.20} \\
        0.3  & 88.98 & 58.33 & 36.06 & 61.12 \\
        0.5  & 88.96 & 58.03 & 35.85 & 60.95 \\
        0.7  & 88.74 & 57.95 & 35.60 & 60.76 \\
        1.0  & 88.26 & 57.38 & 35.35 & 60.33 \\
        \bottomrule
    \end{tabular}
\end{table}

Moreover, increasing $n$ beyond a certain point introduces diminishing returns because additional channel pairs often contribute redundant or low-variance information. In high-dimensional spaces, features are inherently structured such that a few dominant channels encode the majority of the relevant information, while others serve auxiliary roles or capture noise. Randomly selected channel pairs, even in small numbers, are likely to intersect with these dominant channels due to the distributional sparsity of critical features. This explains why the observed success rates remain consistent even with randomly chosen channel pairs, as the adjustments tend to influence significant regions of the feature space regardless of the specific pairs selected.

These findings show two important conclusions. First, the effectiveness of our gradient adjustment method depends more on its ability to mitigate redundancy in a targeted manner than on the exact number of selected channel pairs. Second, the method's primary impact stems from its application itself rather than the specific channels being adjusted, with the critical distinction being between using this method and not using it at all. This robustness to the number of selected pairs underscores the generalizability and efficiency of our approach in optimizing perturbation transferability.

\subsection{Impact of $\tau_{\text{MI}}$ and $\tau_{\text{corr}}$ in Mutual Information (MI) and Pearson correlation (PC)}

Although we adopted an adaptive threshold setting method, we evaluated different configurations of these thresholds to obtain more intuitive information.

The results, presented in Table~\ref{tab:ablation_hyperparams}, indicate that the selected threshold values of 0.5 for mutual information and 0.7 for correlation achieve the highest success rate across ViTs, CNNs, and adversarially trained CNNs. Lower or higher threshold values lead to decreased effectiveness. These findings provide valuable insights for our subsequent, more fine-grained research. 

\begin{table}[ht]
    \caption{Attack success rates (\%) across different step size constants (\(\lambda\)) on ViTs, CNNs, and adversarially trained CNNs. The best results are achieved with \(\lambda = 1.5\).}
    \label{tab:step_size}
    \vskip 0.15in
    \small
    \centering
    \begin{tabular}{c|cccc}
        \toprule
        $\lambda$ & ViTs & CNNs & CNNs-adv & Average \\
        \midrule
        0.5 & 31.72 & 18.83 & 9.60 & 20.05 \\
        0.75 & 56.24 & 31.6 & 15.42 & 34.42 \\
        1.0 & 67.50 & 40.83 & 19.82 & 42.72 \\
        1.25 & 68.70 & 42.05 & 21.15 & 43.97 \\
        1.5 & 69.90 & \textbf{42.70} & \textbf{21.30} & 44.63 \\
        1.75 & 70.38 & 42.6 & 21.22 & \textbf{44.73} \\
        2.0 & \textbf{70.46} & 42.08 & 21.15 & 44.56 \\
        \bottomrule
    \end{tabular}
\end{table}

\begin{figure}[ht]
    \centering
    \begin{tikzpicture}
    \begin{axis}[
        xlabel={Step Size Constant (\(\lambda\))},
        ylabel={Attack Success Rate (\%)},
        xtick={0.5, 0.75, 1.0, 1.25, 1.5, 1.75, 2.0},
        grid=both,
        legend style={font=\small, at={(0.5,1.05)}, anchor=south, legend columns=-1}, 
        ymin=0, ymax=72,
        ytick={10, 20, 30, 40, 50, 60, 70},
        ylabel near ticks
    ]

    \addplot[
        color=blue,
        mark=*,
        thick
    ] coordinates {
        (0.5, 31.72)
        (0.75, 56.24)
        (1.0, 67.50)
        (1.25, 68.70)
        (1.5, 69.90)
        (1.75, 70.38)
        (2.0, 70.46)
    };
    \addlegendentry{ViTs}

    \addplot[
        color=red,
        mark=*,
        thick
    ] coordinates {
        (0.5, 18.83)
        (0.75, 31.60)
        (1.0, 40.83)
        (1.25, 42.05)
        (1.5, 42.70)
        (1.75, 42.60)
        (2.0, 42.08)
    };
    \addlegendentry{CNNs}

    \addplot[
        color=green,
        mark=*,
        thick
    ] coordinates {
        (0.5, 9.60)
        (0.75, 15.42)
        (1.0, 19.82)
        (1.25, 21.15)
        (1.5, 21.30)
        (1.75, 21.22)
        (2.0, 21.15)
    };
    \addlegendentry{CNNs-adv}

    \addplot[
        color=black,
        mark=*,
        thick
    ] coordinates {
        (0.5, 20.05)
        (0.75, 34.42)
        (1.0, 42.72)
        (1.25, 43.97)
        (1.5, 44.63)
        (1.75, 44.73)
        (2.0, 44.56)
    };
    \addlegendentry{Average}

    \end{axis}
    \end{tikzpicture}
    \caption{Attack success rates (\%) for different step size constants (\(\lambda\)) on ViTs, CNNs, adversarially trained CNNs, and their average.}
    \label{fig:step_size_plot}
\end{figure}
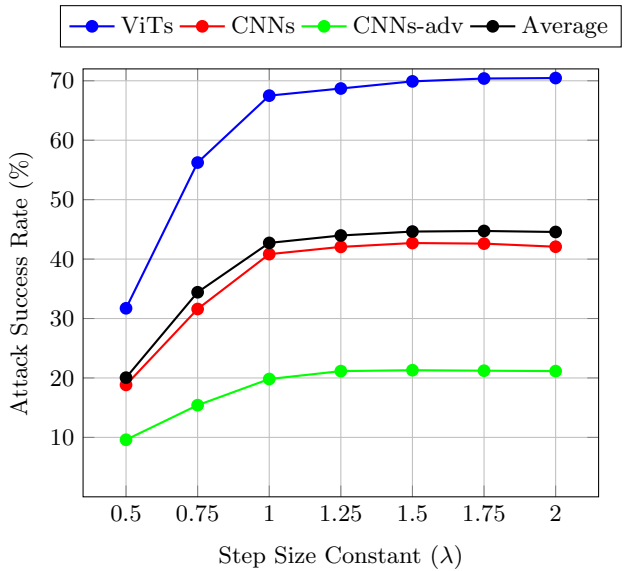

\subsection{Impact of Additional Knowledge Gradient Suppression Factor}

To further investigate the impact of suppressing the gradient of additional knowledge features in ViT variants, we conducted experiments by varying the scaling factor \( c \) applied to the additional knowledge gradient. The scaling factors were set to \( c = \{0.0, 0.1, 0.3, 0.5, 0.7, 1.0\} \), and the resulting attack success rates were evaluated on ViTs, CNNs, and adversarially trained CNNs. 

Table~\ref{tab:distill_token_scaling} presents the attack success rates across different scaling factors. The results demonstrate that \( c = 0.1 \) achieves the best average attack success rate (61.20\%) among all tested settings. This finding demonstrates that \( c = 0.1 \) not only enhances the attack success rate across different model types but also achieves a balanced suppression of the additional knowledge gradient. In contrast, completely nullifying the additional knowledge gradient (\( c = 0.0 \)) or leaving it unadjusted (\( c = 1.0 \)) results in lower success rates, as these extremes either disregard the auxiliary knowledge provided by the additional features or overemphasize it, reducing the transferability of the perturbations. These results prove the importance of carefully tuning the additional knowledge gradient's contribution to optimize adversarial attack performance.

\subsection{Impact of Step Size Constant}

To evaluate the effect of the step size constant \(\lambda\) on attack success rates, we conducted experiments with varying \(\lambda\) values: 0.5, 0.75, 1.0, 1.25, 1.5, 1.75, and 2.0. The objective was to identify the optimal balance between perturbation strength and transferability for adversarial attacks.

The results in Table~\ref{tab:step_size} and Figure~\ref{fig:step_size_plot} show a clear trend where smaller step sizes (\(\lambda < 1.0\)) result in weaker perturbations, leading to lower attack success rates. As \(\lambda\) increases to moderate values (\(1.0 \leq \lambda \leq 1.5\)), the success rates improve significantly across all models, indicating that the perturbations are becoming more effective. This improvement can be attributed to the fact that appropriately increasing the step size introduces a degree of randomness to the generated perturbations, enhancing their generalization and reducing excessive reliance on precise gradient information.

However, for larger step sizes (\(\lambda > 1.5\)), the performance plateaus or slightly declines, particularly for CNNs and adversarially trained CNNs. This suggests that over-amplifying the step size causes the gradient information to lose its guiding role in determining the perturbation direction, ultimately reducing the transferability of the perturbations. This observation underscores the importance of using a step size that is neither too conservative nor overly aggressive.

These findings prove that moderate step sizes, such as \(\lambda = 1.5\), are optimal to achieve a balance between perturbation strength and transferability. By ensuring sufficient perturbation strength without over-amplification, this step size allows for effective attacks across different model architectures while maintaining robustness and generalization.

\end{document}